\documentclass[10pt,onecolumn,letterpaper]{article}
\usepackage[pagenumbers]{cvpr}

\usepackage{amsmath,amsfonts,bm}

\def\1{\bm{1}}

\DeclareMathAlphabet{\mathsfit}{\encodingdefault}{\sfdefault}{m}{sl}
\SetMathAlphabet{\mathsfit}{bold}{\encodingdefault}{\sfdefault}{bx}{n}

\DeclareMathOperator*{\argmin}{arg\,min}

\usepackage{amsthm}
\usepackage{multirow}
\usepackage{algorithm}
\usepackage{algorithmic}

\newcommand{\Engram}{\textsc{Engram}}

\newcommand{\real}{\mathbb{R}}
\newcommand{\Exp}{\mathbb{E}}
\newcommand{\Varr}{\operatorname{Var}}
\newcommand{\Covv}{\operatorname{Cov}}
\newcommand{\ff}{\mathbf{f}}
\newcommand{\pp}{\mathbf{p}}
\newcommand{\bfeps}{\boldsymbol{\varepsilon}}

\usepackage{float}
\definecolor{cvprblue}{rgb}{0.21,0.49,0.74}
\usepackage[pagebackref,breaklinks,colorlinks,allcolors=cvprblue]{hyperref}
\title{\Engram{}: Reliability-Weighted Latent Fusion and Occlusion-Aware Layered Readout for Video World Models}
\author{\parbox{0.97\textwidth}{\centering
  \normalsize
  Chenhao Wu\textsuperscript{1,*}\quad
  Zhaolong Su\textsuperscript{2,*}\quad
  Yang Liu\textsuperscript{3,*}\quad
  Tianrui Su\textsuperscript{4}\\[2pt]
  Yingguang Yang\textsuperscript{3}\quad
  Shang Luo\textsuperscript{3}\quad
  Yancheng Chen\textsuperscript{1}\quad
  Chongyang Zhang\textsuperscript{5}\\[2pt]
  Hao Zheng\textsuperscript{5}\quad
  Kefu Xu\textsuperscript{3}\quad
  Congjing Ran\textsuperscript{6}\quad
  Bin Chong\textsuperscript{3,\ensuremath{\dagger}}\\[5pt]
  \small
  \textsuperscript{1}University of the Chinese Academy of Sciences\quad
  \textsuperscript{2}Jilin University\quad
  \textsuperscript{3}Peking University\\
  \textsuperscript{4}University of New South Wales\quad
  \textsuperscript{5}Fullive-AI\quad
  \textsuperscript{6}Wuhan University\\[3pt]
  \footnotesize
  \textsuperscript{*}Equal contribution.\quad
  \textsuperscript{\ensuremath{\dagger}}Corresponding author.
}}

\makeatletter
\g@addto@macro\@maketitle{%
  \noindent\begin{minipage}{\textwidth}
    \centering
    \includegraphics[width=0.92\linewidth]{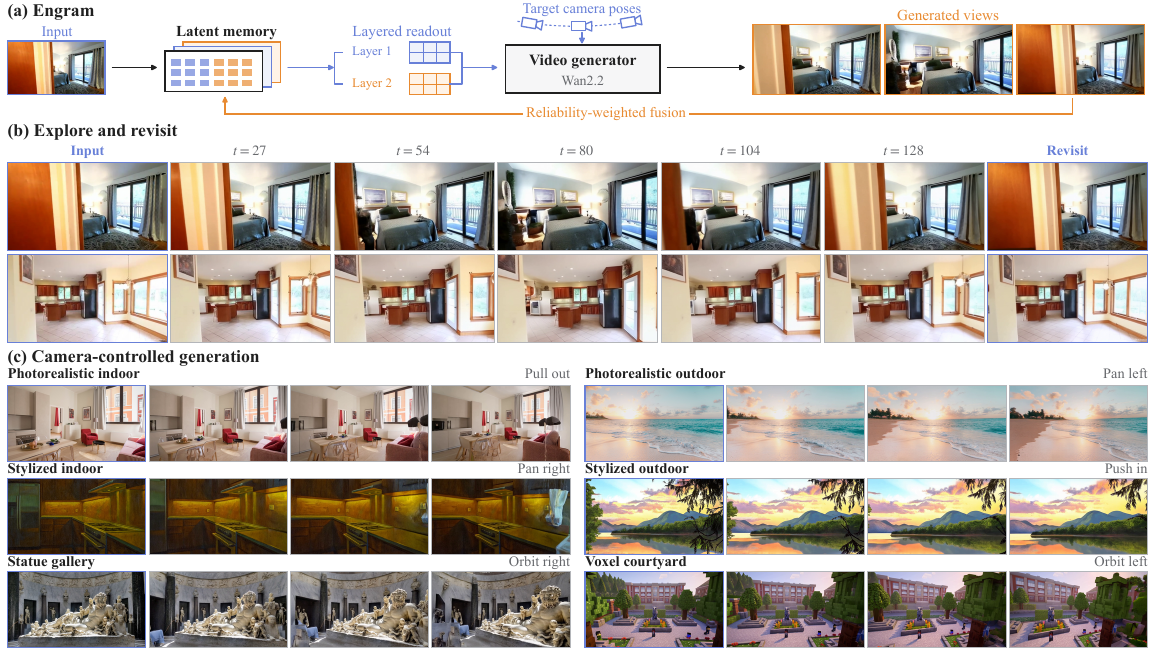}
    \captionsetup{hypcap=false}
    \captionof{figure}{
    \textbf{Explore, remember, and revisit with \Engram{}.}
    (a) Persistent latent memory with layered readout and reliability-weighted writeback.
    (b) Closed-loop revisits on RealEstate10K.
    (c) Camera-controlled generation across diverse WorldScore scenes.
}
\label{fig:teaser}

  \end{minipage}\par
  \vspace{4pt}
}
\makeatother

\makeatletter
\patchcmd{\@maketitle}{\vskip .375in}{\vskip 0pt}{}{\errmessage{Title top-spacing patch failed}}
\patchcmd{\@maketitle}{\vspace*{24pt}}{\vspace*{12pt}}{}{\errmessage{Title author-spacing patch failed}}
\patchcmd{\@maketitle}{\vspace*{12pt}}{\vspace*{4pt}}{}{\errmessage{Title spacing patch failed}}
\patchcmd{\@maketitle}{width=0.92\linewidth}{width=0.88\linewidth}{}{\errmessage{Teaser sizing patch failed}}
\patchcmd{\@maketitle}{\vspace*{12pt}}{\vspace*{0pt}}{}{\errmessage{Title footer-spacing patch failed}}
\patchcmd{\@maketitle}{\vskip .5em}{\vskip 0pt}{}{\errmessage{Author footer-spacing patch failed}}
\patchcmd{\@maketitle}{\vspace{4pt}}{\vspace{0pt}}{}{\errmessage{Teaser footer-spacing patch failed}}
\patchcmd{\abstract}{\vspace*{12pt}}{\vspace*{6pt}}{}{\errmessage{Abstract spacing patch failed}}
\makeatother

\begin{document}
\raggedbottom
\maketitle
\begin{abstract}
Camera-conditioned video world models aim to generate 3D-consistent videos from a single image along specified camera trajectories. However, set-union updates in latent spatial memory leave repeated observations unconsolidated, while frontmost-only readout limits simultaneous access to stored evidence at different depths. 
To address these limitations, we present \Engram{}, a latent spatial memory framework combining \emph{reliability-weighted latent fusion} and \emph{occlusion-aware layered readout}. Specifically, recursive fusion integrates geometrically and visually compatible observations in a channel-normalized latent space, consolidating repeated evidence into a shared scene representation. Layered readout exposes multiple depth-ordered latent candidates with their reliability and availability at each target cell. This gives the generator access to information from multiple surfaces as visibility changes. Explicit camera rays provide viewpoint information independently of memory coverage, while geometric consistency constraints guide memory updates. 
Memory retrieval uses a single projection at latent resolution, avoiding intermediate RGB rendering and VAE re-encoding. We adapt a pretrained video generator using only 256 training scenes. On 100 RealEstate10K test scenes, \Engram{} improves novel-view-synthesis and closed-loop PSNRs over Spatia by 0.82 and 4.48\,dB, respectively. On a 200-case WorldScore evaluation subset, it achieves an average score of 67.78, exceeding Spatia by 3.31 points.
Project page: \url{https://wchkoutian.github.io/engram-website/}.
\end{abstract}

\section{Introduction}
\label{sec:intro}
Large-scale video diffusion models synthesize photorealistic sequences and provide a foundation for generative world models~\citep{wan2025,yang2025cogvideox,hacohen2024ltxvideo}. Geometric drift is what separates a world model from a per-frame generator: driven along a camera trajectory, a model produces frames that are individually plausible and jointly inconsistent, and the inconsistency compounds from chunk to chunk. Explicit spatial memory removes the drift by preserving observed geometry and appearance and by showing each newly generated chunk a view of that memory from the target camera.

Existing systems keep explicit 3D representations, including colored point clouds~\citep{huang2025voyager,yu2024viewcrafter} and Gaussian scene representations~\citep{yu2025wonderworld,chen2025flexworld}, render them at every step, and pass the render back into the conditioning path through a variational autoencoder (VAE)~\citep{huang2025voyager,yu2024viewcrafter,chen2025flexworld}, paying for the render and for a second lossy encode. LSM-World~\citep{wang2026lsmworld} removes the intermediate by storing scene features in the diffusion latent space, lifting latent tokens to 3D with depth-guided back-projection and reading them back with a projection at latent resolution. We retain this direct latent-space projection. Removing the render, however, does not decide what such a store should hold. \emph{What} does a slot represent once the same surface has been observed many times? And \emph{which} stored slot should the generator be shown when several surfaces project to the same target cell?

Set-union accumulation, the natural update for such a store, makes every valid back-projection its own record. A surface seen across a hundred overlapping views therefore occupies a hundred competing entries whose features differ only by depth noise and per-view estimation error. Nothing in the store decides which of them is right, so the decision falls to readout, and readout is where it fails: a small change in estimated depth moves the choice to a different record of the same surface, and a revisit renders that surface inconsistently with the visit that already recorded it. The failure is structural rather than occasional, because overlapping views are precisely what a memory is built to collect. Fusion resolves the competition where it can still be resolved, at write time against a shared scene representation, instead of deferring it to every query. It must also decide what not to store: a generated record can project in front of an initial surface that it never observed, and writing it there destroys the only evidence of that surface.

\begin{figure}[t]
    \centering
    \includegraphics[width=0.62\linewidth]
        {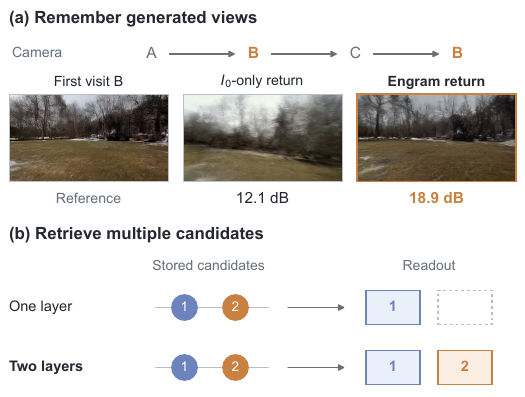}
    \caption{
        \textbf{Retaining and retrieving scene history.}
        (a) Retaining generated history improves consistency
        when revisiting a previously generated view.
        PSNR is measured against the first generated visit
        in one development-set example.
        (b) Layered readout exposes multiple stored candidates.
        This schematic does not attribute the improvement
        in (a) specifically to layered readout.
    }
    \label{fig:engram_motivation}
\end{figure}

A second difficulty concerns which records the generator can address. Frontmost selection exposes only the nearest surface at each target cell, leaving other projected candidates unavailable to the generator in that readout. A revisit at a shallower angle, or a ray that grazes an occlusion boundary, exposes the gap, and a single-layer readout has no place to put the answer.
Figure~\ref{fig:engram_motivation} illustrates two requirements:
retaining generated history for revisits and exposing multiple
stored candidates to the generator.

A third challenge is learning spatial-memory conditioning.
Training a generator from scratch would require learning both video
synthesis and memory use. Our two-stage adaptation reuses pretrained
synthesis capabilities, first learning the memory interface and then
adapting the generator. Spatia also adapts a pretrained generator,
using 50,000 training videos~\citep{zhao2026spatia}.
Counting optimization steps times effective batch size, our schedule
uses 48,000 sample presentations versus 832,000 calculated from
Spatia's reported schedule: 5.8\% as many, or a 17.3-fold difference
in this count. Supplementary Sec.~\ref{sec:training_accounting}
gives the calculation and counting convention.

We introduce \Engram{}, a latent spatial memory that resolves both questions inside the read--generate--update cycle illustrated in Figure~\ref{fig:teaser}. \emph{Reliability-weighted latent fusion} consolidates compatible observations within a voxel into a single slot, weighted by depth reliability and gated so that a generated record cannot enter in front of an initial surface it never observed. \emph{Occlusion-aware layered readout} returns depth-ordered candidates per cell together with their reliability and availability, so the answer stays addressable after the nearest surface stops being the right one. Explicit \emph{Pl\"ucker camera rays} carry camera information independently of memory coverage. Generated frames are written back through the same cycle, so a view survives its departure from the context window and is still available when the camera returns. We adapt a pretrained Wan2.2 backbone using 256 training scenes. We evaluate \Engram{} on RealEstate10K~\citep{zhou2018stereo} and WorldScore~\citep{duan2025worldscore}, covering novel-view synthesis, closed-loop revisits, and camera-controlled world generation.

\textbf{Contributions.}
(i) We introduce a \emph{reliability-weighted latent fusion} operator that consolidates compatible multi-view observations of a surface into a single slot, weighted by depth reliability and gated against writing a generated record in front of an initial surface. The store stops accumulating competing records, and readout stops having to adjudicate among them.
(ii) We propose an \emph{occlusion-aware layered readout} that returns multiple depth-ordered candidates per cell together with their reliability and availability, so that evidence behind the nearest surface remains addressable to the generator as visibility changes.
(iii) We demonstrate \emph{data-efficient spatial-memory adaptation}
through a two-stage strategy that learns the memory interface before
adapting the generator, enabling consistent generation with limited
training data.

\section{Related Work}
\label{sec:related}

\paragraph{Video diffusion world models.}
Video diffusion transformers~\citep{peebles2023dit,wan2025,yang2025cogvideox,hacohen2024ltxvideo} generate temporally coherent video and are increasingly treated as world models when conditioned on a camera trajectory. Without a persistent representation they drift geometrically over long rollouts, which motivates explicit spatial memory.

\paragraph{Spatial memory for consistent generation.}
Existing methods maintain explicit 3D representations and render
target views for generation~\citep{yu2025wonderworld,
yu2024wonderjourney,huang2025voyager,yu2024viewcrafter,
chen2025flexworld,engstler2025invisiblestitch,chung2023luciddreamer},
while view-memory methods retrieve past frames~\citep{seva2025,vmem2025}.
Volumetric and learned latent fusion have been studied for
reconstruction~\citep{curless1996volumetric,newcombe2011kinectfusion,
weder2021neuralfusion,huang2021difusion}.
Our closest baseline, LSM-World~\citep{wang2026lsmworld},
stores diffusion features in 3D using set-union updates and
frontmost readout.
We retain direct latent-space projection and introduce
reliability-weighted latent fusion and occlusion-aware layered
readout within the generation loop.

\paragraph{Camera conditioning and calibration.}
Pl\"ucker coordinates~\citep{plucker1865} are used for camera conditioning~\citep{he2024cameractrl}. We use these coordinates to provide viewpoint information independently of memory coverage. Confidence calibration aligns classification probabilities with empirical correctness~\citep{guo2017calibration}. Our memory fusion uses geometric and temporal precision proxies to weight observations.

\section{Preliminaries}
\label{sec:prelim}

\paragraph{Chunked autoregressive video world models.}
Given an initial image $I_0$ and a camera trajectory $\{E_k\}_{k=0}^{N}$, a video world model generates a sequence $\{I_k\}_{k=0}^{N}$ consistent with the prescribed viewpoints. Here, $E_k\in\mathrm{SE}(3)$ maps world coordinates to camera coordinates, and $K_l$ denotes the camera intrinsics at latent resolution, with the frame index omitted for simplicity. Generation proceeds in temporally overlapping chunks; we index chunks by $t$ and reserve $k$ for frames, so that $E_{t,j}$ is the $j$-th camera of chunk $t$. For each chunk, the model queries a persistent spatial memory $\mathcal{M}$ from the target viewpoints, generates the corresponding video, and integrates new observations into the memory.

\paragraph{Backbone and conditional generation.}
We use Wan2.2-TI2V-5B~\citep{wan2025} as the pretrained video generation backbone. Its VAE has a temporal compression factor of four and $C=48$ latent channels. The VAE downsamples each spatial dimension by a factor of 16. A generated chunk comprises several latent frames $Z_{t,j}$, indexed by the same $j$ as its cameras. We optimize a conditional velocity field using flow matching~\citep{lipman2023flowmatching}. For clean target latents $x$ and noise $\varepsilon\sim\mathcal{N}(0,I)$, the interpolation is $x_\sigma=(1-\sigma)x+\sigma\varepsilon$, where $\sigma\in[0,1]$. The training objective is
\begin{equation}
\mathcal{L}_{\mathrm{FM}}
=
\mathbb{E}_{\sigma,x,\varepsilon}
\left[
\left\|
v_\theta(x_\sigma,\sigma,c)-(\varepsilon-x)
\right\|_{\Omega}^{2}
\right],
\label{eq:fm}
\end{equation}
where $\Omega$ contains the valid prediction targets and $\|\cdot\|_{\Omega}^{2}$ denotes their mean squared error. The conditioning $c$ comprises spatial-memory features, camera information, a clean prefix, and text when provided. The prefix is retained as unperturbed context. Spatial conditioning enters through a ControlNet-style side branch~\citep{zhang2023controlnet,jiang2025vace}.

\paragraph{Latent spatial memory and baseline operators.}
Let $z\in\real^{C\times H_l\times W_l}$ denote a latent frame, with depth $D\in\real^{H_l\times W_l}$, whose entries we write $D_{uv}$, and intrinsics $K_l$ at the same resolution. A valid latent cell $(u,v)$ of a frame captured by a camera $E$ is lifted to a world point $\pp_{uv}=E^{-1}\pi^{-1}(u,v,D_{uv};K_l)$ and associated with its feature $\ff_{uv}=z_{:,v,u}$, where $\pi^{-1}$ denotes camera back-projection. LSM-World~\citep{wang2026lsmworld} accumulates these observations through set union,
\begin{equation}
\mathcal{M}
\leftarrow
\mathcal{M}\cup
\{(\pp_{uv},\ff_{uv}):(u,v)\in\Lambda\},
\label{eq:union}
\end{equation}
where $\Lambda$ is the set of valid cells admitted into the baseline memory. For a target camera $E$, let $\mathcal{P}(u,v)$ contain the cached points that project to cell $(u,v)$ with positive camera depth. For nonempty $\mathcal{P}(u,v)$, the readout selects the frontmost feature,
\begin{equation}
\hat{\ff}_{u,v}=\ff_{j^\star},
\qquad
j^\star=
\argmin_{j\in\mathcal{P}(u,v)} d_{j},
\label{eq:argmin}
\end{equation}
where $d_{j}$ is the depth of point $j$ in the target camera. A validity mask identifies cells without projected observations. \Engram{} replaces set-union accumulation and frontmost selection with reliability-weighted latent fusion and occlusion-aware layered readout, while retaining direct projection at latent resolution.

\paragraph{Notation.}
For \Engram{}, $w_i\ge0$ denotes the reliability weight of observation $i$, $W_a\ge0$ the accumulated weight of memory slot $a$, and $\gamma\in(0,1]$ the temporal discount base for generated observations. Slots are grouped by the voxel index $\mathbf{k}=\lfloor\pp/h_{\mathrm{vox}}\rfloor$, where $h_{\mathrm{vox}}$ is expressed in the same scene-normalized coordinates as $\pp$. Each voxel holds a bounded number of slots, at most $L$, which may correspond to different depths or different provenance, while the occupied voxel set expands as new regions are observed. We use $(u,v)$ exclusively for latent-cell coordinates.

\section{The \Engram{} Model}
\label{sec:method}

\subsection{Overview}
\label{sec:overview}

\begin{figure*}[t]
    \centering
    \includegraphics[width=\textwidth]{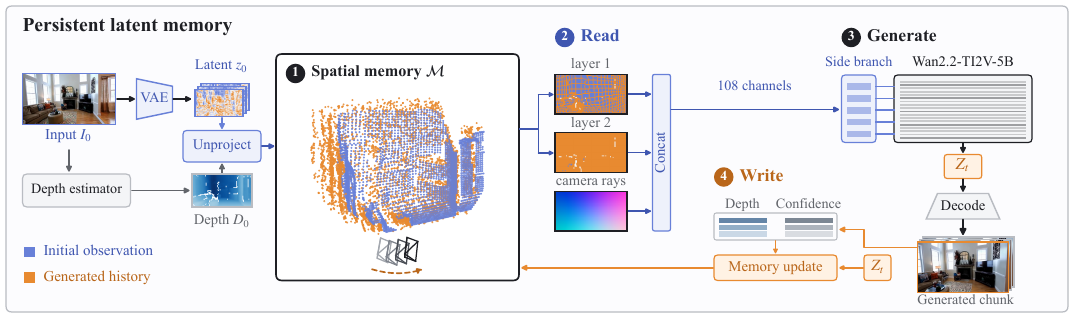}
    \caption{
        \textbf{Overview of \Engram{}.}
        The input image initializes spatial latent memory $\mathcal{M}$.
        Two readout layers and camera rays condition the video generator
        through a side branch. Generated latents are written back using
        geometry estimated from decoded frames; matching $Z_t$ ports denote
        the same output tensor. Blue indicates initial observations and
        orange indicates generated history. The cloud is subsampled from
        a recorded memory, and the layer maps are recomputed at a target pose.
        Module and depth/confidence glyphs are schematic.
    }
    \label{fig:overview}
\end{figure*}

\begin{figure*}[t]
    \centering
    \includegraphics[width=0.888\textwidth]{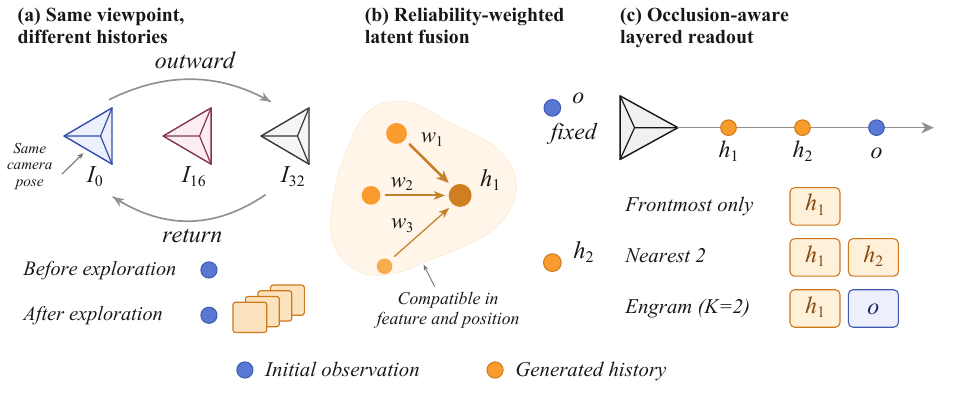}
    \caption{
        \textbf{Reliability-weighted latent fusion and occlusion-aware layered readout.}
        (a) Revisiting the same viewpoint after exploration provides
        additional observations.
        (b) Compatible generated observations are fused using
        reliability weights, while the initial observation is preserved.
        (c) Layered readout exposes depth-ordered candidates and retains
        initial-view evidence when available.
    }
    \label{fig:memory_operators}
\end{figure*}

\Engram{} initializes a persistent latent spatial memory and a clean prefix from the input image. 
It then generates successive video chunks through a read--generate--update cycle (Algorithm~\ref{alg:rollout}). 
At each chunk, the memory provides depth-ordered scene features for the target viewpoints, while explicit camera rays supply viewpoint information independently of memory coverage.
The conditioned video generator synthesizes the next chunk, whose latent observations are integrated into memory using estimated geometry and reliability weights.
The two central operators are reliability-weighted latent fusion and occlusion-aware layered readout, supported by geometric consistency constraints and a two-stage adaptation strategy.
Figure~\ref{fig:overview} shows the full cycle, while
Fig.~\ref{fig:memory_operators} details the two operators.

\begin{algorithm*}[t]
\caption{\Engram{} rollout. Initialization runs once, before the first
chunk; the remaining steps run once per generation chunk. The memory
$\mathcal{M}$ and the clean prefix $z_{\mathrm{prefix}}$ persist across
chunks.}
\label{alg:rollout}
\begin{algorithmic}[1]
\REQUIRE Input image $I_0$ and its camera $E_0$, target cameras
$\mathcal{E}_t=\{E_{t,j}\}$ for the current chunk $t$,
latent-resolution intrinsics $K_l$, and readout depth $K$.

\STATE \textbf{Initialize (once):}
estimate the depth map $D_0$ of $I_0$, back-project its valid pixels
through $(E_0,K_l)$ to seed the spatial memory $\mathcal{M}$ with
initial observations, and encode $I_0$ to obtain the clean prefix
$z_{\mathrm{prefix}}$.

\STATE \textbf{Read:}
$R_t \leftarrow
\textsc{LayeredReadout}(\mathcal{M},\mathcal{E}_t,K_l,K)$.
\hfill Sec.~\ref{sec:readout}

\STATE \textbf{Camera:}
construct Pl\"ucker rays $\boldsymbol{\ell}_t$
relative to $E_{t,0}$ and form the spatial condition
$c_t^{\mathrm{sp}}=[R_t;\boldsymbol{\ell}_t]$.
\hfill Sec.~\ref{sec:camera}

\STATE \textbf{Generate:}
$Z_t \leftarrow
\textsc{Generate}(c_t^{\mathrm{sp}},z_{\mathrm{prefix}},
\text{text})$.

\STATE \textbf{Decode and estimate geometry:}
$X_t \leftarrow \mathcal{D}(Z_t)$;
estimate depth maps at the timestamps of the latent frames.
\hfill Sec.~\ref{sec:update}

\FORALL{new latent frames $j$ in the chunk}
    \STATE Back-project valid observations from
    $(Z_{t,j},D_{t,j},E_{t,j},K_l)$.
    \STATE Compute observation weights using
    Eq.~\eqref{eq:reliability}.
    \STATE Apply the geometric admission rule.
    \hfill Sec.~\ref{sec:update}
    \STATE Aggregate accepted observations within each voxel
    and match candidates to compatible slots.
    \hfill Sec.~\ref{sec:fusion}
    \STATE Fuse matched candidates using Eq.~\eqref{eq:fuse}
    and allocate slots for unmatched candidates when capacity permits.
\ENDFOR

\STATE \textbf{Handoff:}
refine the encoding of the last generated frame
to obtain the next clean prefix.
\hfill Eq.~\eqref{eq:handoff}

\end{algorithmic}
\end{algorithm*}

\subsection{Reliability-weighted latent fusion}
\label{sec:fusion}

\paragraph{Observation reliability.}
For each generated observation $i$, we estimate a unit surface
normal $\mathbf{n}_i$ from finite differences of the back-projected
depth and obtain its unit viewing direction $\mathbf{d}_i$.
Both vectors are expressed in the same camera coordinate system.
We define
\begin{equation}
g_i=
\operatorname{clip}
\bigl(|\mathbf{n}_i^\top\mathbf{d}_i|,\epsilon,1\bigr),
\qquad
w_i=\sqrt{g_i}\,\gamma^{r_i},
\label{eq:reliability}
\end{equation}
where $\epsilon>0$ is a small positive floor and $r_i$ is the
observation's temporal index at latent resolution, measured from
the start of the rollout.
The angular factor down-weights grazing observations, while
the temporal factor moderates the contribution of later
generated observations.
Both factors are precision proxies rather than fitted error
variances: $g_i$ is the grazing-angle precision of the observation,
computed from the same normal estimate, and $\gamma^{r_i}$ encodes the
known autoregressive bias of generated frames.

\paragraph{Feature normalization.}
We normalize latent features using fixed channel statistics $(\boldsymbol{\mu},\Sigma)$, where $\Sigma$ is a positive diagonal matrix of channel variances with numerical stabilization:
\begin{equation}
\tilde{\ff}_i
=
\Sigma^{-1/2}(\ff_i-\boldsymbol{\mu}).
\label{eq:whiten}
\end{equation}
This normalization places latent channels on comparable scales for appearance compatibility testing. The statistics remain fixed throughout memory construction and retrieval. During readout, we restore the original latent coordinates through $\ff=\Sigma^{1/2}\tilde{\ff}+\boldsymbol{\mu}$.

Before fusion, we aggregate accepted observations within each voxel of the current view using their reliability weights. In the update below, $\bar{\ff}_i$ and $\bar{\pp}_i$ denote the aggregated feature and position of that voxel, and $\bar{w}_i$ its summed weight. We match each candidate to the closest compatible slot in feature space within the same voxel. Compatibility requires
$\|\bar{\ff}_i-\tilde{\ff}_a\|_2<\tau_{\mathrm f}$ and $\|\bar{\pp}_i-\pp_a\|_2/h_{\mathrm{vox}}<\tau_{\mathrm p}$.
Generated candidates are matched only to slots formed from generated observations, preserving the initial observation. For a matched pair, we apply
\begin{equation}
\begin{aligned}
\tilde{\ff}_a
&\leftarrow
\frac{W_a\tilde{\ff}_a+\bar{w}_i\bar{\ff}_i}{W_a+\bar{w}_i},\\
W_a
&\leftarrow
\min(W_a+\bar{w}_i,W_{\max}).
\end{aligned}
\label{eq:fuse}
\end{equation}
The weight cap limits the accumulated influence of earlier observations. An unmatched candidate initializes a new slot with weight $\min(\bar w_i,W_{\max})$, subject to a per-voxel limit of $L$ slots. Unmatched candidates are discarded when the voxel has reached capacity.

\subsection{Occlusion-aware layered readout}
\label{sec:readout}

For each target view, we project memory slots with positive camera depth onto the latent grid and order them by depth within each cell.
We first select up to $K$ frontmost slots. If none originates from the initial observation, the nearest projected slot from that observation replaces the last selected slot, when available.

Each selected layer provides a latent feature ($C$ channels), a bounded reliability value, a validity mask, and a normalized inverse depth.
Features are restored to their original latent coordinates before conditioning, and missing layers are filled with zeros and marked invalid.
The resulting memory conditioning has $K(C+3)$ channels, giving $102$ channels for $C=48$ and $K=2$. This layered representation provides the generator with stored information at multiple depths as visibility changes.

\subsection{Explicit Pl\"ucker camera conditioning}
\label{sec:camera}

Explicit camera rays provide a dense viewpoint signal independently of memory coverage.
For each target camera $E_{t,j}$, we use the relative extrinsic $E_{t,j}^{\mathrm{rel}}=E_{t,j}E_{t,0}^{-1}$, where $E_{t,0}$ is the first camera of the current chunk.
For each latent cell $(u,v)$, we compute the unit ray direction $\mathbf{r}_{uv}$ and camera center $\mathbf{o}$ in this reference camera coordinate system.
The Pl\"ucker representation is
\begin{equation}
\boldsymbol{\ell}_{uv}
=
\bigl(\mathbf{r}_{uv},\mathbf{m}_{uv}\bigr),
\qquad
\mathbf{m}_{uv}
=
\mathbf{o}\times\mathbf{r}_{uv},
\label{eq:plucker}
\end{equation}
following~\citep{plucker1865,he2024cameractrl}. The six ray channels are concatenated with the memory readout and injected through the conditioning branch.
For $C=48$ and $K=2$, the combined spatial conditioning contains $108$ channels.
Cells without projected memory therefore retain explicit information about the target viewpoint.

\subsection{Continuous generation and memory update}
\label{sec:update}

After generation, we decode the chunk and estimate depth at the timestamps of its latent frames. The estimated geometry is aligned with the latent grid and memory coordinate system, then paired with the original generated latent features. Only observations with finite, positive depth are eligible for writing. The geometric admission rule filters generated candidates against the initial-view depth before memory insertion. Accepted observations are integrated through the fusion procedure in Sec.~\ref{sec:fusion}.

At each chunk boundary, we encode the last generated frame $I_{\mathrm{last}}$ and refine its latent through the fixed VAE decoder:
\begin{equation}
z_{\mathrm{prefix}}^\star
\approx
\argmin_z
\left\|
\mathcal{D}(z)-I_{\mathrm{last}}
\right\|_1.
\label{eq:handoff}
\end{equation}
The refined latent serves as the clean prefix for the next generation chunk.

\subsection{Training and implementation}
\label{sec:training}

We adapt the pretrained generator in two stages. Stage~1 trains the conditioning branch while keeping the generation backbone and VAE fixed. Stage~2 freezes the conditioning branch and optimizes all parameters of the generation Transformer. The VAE remains fixed throughout. Both stages use the flow-matching objective in Eq.~\eqref{eq:fm}, with the clean prefix retained as unperturbed context and excluded from the prediction loss.

Training uses 256 scenes, each providing five forward windows and four return windows, yielding 2,304 training windows. Training memories are constructed from observed video history using precomputed geometry. Each window includes a text caption, which is dropped with probability $0.1$ during training.

The two stages run for 4,000 and 8,000 optimization steps, respectively, with an effective batch size of four across four NVIDIA H200 GPUs. We use AdamW, bfloat16 mixed precision, and gradient checkpointing. At inference, depth is estimated from decoded frames using MapAnything~\citep{keetha2026mapanything}, and video chunks are sampled with UniPC~\citep{zhao2023unipc} using 40 steps.

\subsection{Why fusion improves on selection}
\label{sec:theory}

An idealized observation model illustrates the benefit of
combining repeated evidence.
Consider observations of a fixed scene location with independent,
zero-mean latent errors and covariance $\sigma_i^2 I$.
Fixed inverse-variance weighting is unbiased and yields covariance
$\bigl(\sum_i \sigma_i^{-2}\bigr)^{-1}I$, which is no larger
than that of any individual observation, and is therefore also no
worse than selecting any candidate fixed in advance.

Frontmost selection uses the depth-argmin over the candidates.
For repeated observations of the same surface with i.i.d., zero-mean,
non-degenerate depth errors, the minimum observed depth is biased
toward the camera (Appendix~\ref{app:theory}).
Fusion aggregates compatible candidates through a weighted mean.

Our practical operator uses geometric and temporal reliability weights
with compatibility matching and bounded accumulation.
Appendix~\ref{app:theory} analyzes weighted averaging, depth-based
selection, and the effects of correlated errors and observation-dependent weights.

\begin{table*}[!t]
\centering
\caption{\textbf{Evaluation on 200 WorldScore cases.}
Static averages the seven control and quality
metrics; Dynamic also includes the three motion metrics; Avg.\ is their mean.
All scores are higher-is-better. A dash indicates a metric or aggregate
not applicable to the method.
Best and second-best scores are shown in bold and underlined, respectively.}
\label{tab:worldscore}
\footnotesize
\setlength{\tabcolsep}{3pt}
\resizebox{\textwidth}{!}{%
\begin{tabular}{@{}l*{13}{c}@{}}
\toprule
Method
& \multicolumn{3}{c}{Overall}
& \multicolumn{3}{c}{Control and content}
& \multicolumn{4}{c}{Geometry and perception}
& \multicolumn{3}{c}{Motion} \\
\cmidrule(lr){2-4}
\cmidrule(lr){5-7}
\cmidrule(lr){8-11}
\cmidrule(lr){12-14}
& Avg. & Static & Dynamic
& Cam. & Obj. & Content
& 3D & Photo. & Style & Subj.
& Acc. & Mag. & Smooth. \\
\midrule
\multicolumn{14}{@{}l}{\emph{Models with 3D cache}} \\
WonderWorld~\citep{yu2025wonderworld}
& -- & 63.82 & --
& \textbf{67.61} & 40.75 & \textbf{68.78}
& 79.43 & 84.53 & 62.10 & 43.56
& -- & -- & -- \\
Spatia~\citep{zhao2026spatia}
& \underline{64.47} & 66.65 & 62.29
& 55.67 & \textbf{65.75} & 37.75
& 85.83 & 89.49 & 82.09 & 49.96
& \underline{58.27} & 26.48 & 71.64 \\
LSM-World~\citep{wang2026lsmworld}
& 56.59 & 59.70 & 53.47
& 57.27 & 38.25 & 42.64
& 83.59 & 64.41 & 80.58 & 51.17
& 49.08 & 26.74 & 40.94 \\
\midrule
\multicolumn{14}{@{}l}{\emph{General video models}} \\
Wan2.2-TI2V-5B~\citep{wan2025}
& 64.27 & \underline{67.33} & 61.21
& 33.53 & 54.59 & 46.84
& \underline{90.01} & 89.33 & \textbf{95.12} & \textbf{61.88}
& 55.01 & \underline{28.93} & 56.86 \\
LTX-Video~\citep{hacohen2024ltxvideo}
& 63.48 & 64.58 & \underline{62.39}
& 30.82 & \underline{61.25} & 49.34
& \textbf{91.23} & \textbf{94.94} & 72.81 & 51.67
& 57.85 & 27.11 & \underline{86.83} \\
CogVideoX~\citep{yang2025cogvideox}
& 60.34 & 62.66 & 58.02
& 44.80 & 42.00 & \underline{55.09}
& 81.00 & 81.82 & 74.46 & 59.44
& \textbf{59.81} & \textbf{30.70} & 51.13 \\
\midrule
\Engram{} (ours)
& \textbf{67.78} & \textbf{69.97} & \textbf{65.60}
& \underline{66.92} & 57.00 & 41.89
& 85.84 & \underline{91.90} & \underline{86.67} & \underline{59.57}
& 52.47 & 24.76 & \textbf{88.95} \\
\bottomrule
\end{tabular}%
}
\end{table*}

\section{Experiments}
\label{sec:exp}

We evaluate complete-system performance on WorldScore and
RealEstate10K, and isolate the effects of memory writing,
layered readout, and two-stage adaptation through ablations.

\begin{table}[!t]
\centering
\caption{\textbf{Evaluation on 100 RealEstate10K scenes.}
NVS evaluates frames 1--80; closed-loop revisit compares frame 152
with the initial image. All methods use the same evaluation protocol.
Best and second-best scores are bold and underlined, respectively.}
\label{tab:re10k}
\footnotesize
\setlength{\tabcolsep}{3pt}
\resizebox{0.70\linewidth}{!}{%
\begin{tabular}{@{}lcccccc@{}}
\toprule
& \multicolumn{3}{c}{NVS}
& \multicolumn{3}{c}{Closed-loop revisit} \\
\cmidrule(lr){2-4}\cmidrule(lr){5-7}
Method
& PSNR$\uparrow$ & SSIM$\uparrow$ & LPIPS$\downarrow$
& PSNR$\uparrow$ & SSIM$\uparrow$ & LPIPS$\downarrow$ \\
\midrule
ViewCrafter~\citep{yu2024viewcrafter}
& 8.494 & 0.3405 & 0.7852
& 9.540 & 0.3467 & 0.7468 \\
FlexWorld~\citep{chen2025flexworld}
& 11.167 & 0.5805 & 0.6684
& 12.191 & 0.6089 & 0.5612 \\
Spatia~\citep{zhao2026spatia}
& \underline{16.745} & \underline{0.6832} & \underline{0.3525}
& \underline{15.978} & \underline{0.6827} & \underline{0.3373} \\
LSM-World~\citep{wang2026lsmworld}
& 15.624 & 0.6311 & 0.4148
& 12.235 & 0.3716 & 0.7487 \\
\midrule
\Engram{} (ours)
& \textbf{17.568} & \textbf{0.6975} & \textbf{0.3212}
& \textbf{20.455} & \textbf{0.7692} & \textbf{0.2200} \\
\bottomrule
\end{tabular}%
}
\end{table}

\subsection{Main comparisons and shared protocol}
We compare complete systems using released checkpoints, except for
LSM-World~\citep{wang2026lsmworld}, which we reproduce.
Wan2.2-TI2V-5B~\citep{wan2025} serves as a backbone reference
with image and text conditioning rather than pose tensors;
its comparison with \Engram{} is not an isolated memory ablation.

\paragraph{WorldScore.}
We evaluate on a 200-case WorldScore subset using its ten
official metrics~\citep{duan2025worldscore}.
WonderWorld's motion metrics and dependent aggregates are
not applicable.
General video models receive image and camera-text prompts,
whereas pose-conditioned methods receive numeric camera poses.
Table~\ref{tab:worldscore} shows that \Engram{} achieves the highest
aggregate score of 67.78. Relative to Spatia, motion smoothness,
camera control, and subjective quality improve by 17.31, 11.25,
and 9.61 points, respectively. Its 3D consistency is nearly
unchanged, while motion accuracy and magnitude remain lower.
Detailed protocols are provided in Supplementary
Sec.~\ref{sec:eval_protocol}.

{\widowpenalty=10000\brokenpenalty=10000
\paragraph{RealEstate10K.}
We evaluate all methods on the same 100 held-out
scenes~\citep{zhou2018stereo,zhao2026spatia}.
NVS scores frames 1--80 against ground truth; closed-loop
evaluation compares frame 152 with the initial image.
Methods share target camera trajectories, field of view,
frame times, and scoring resolution.
Each task starts from the real input image and uses only
generated observations as subsequent visual history.
Table~\ref{tab:re10k} reports mean per-scene scores.
\par}

Table~\ref{tab:re10k} shows that \Engram{} achieves the best PSNR,
SSIM, and LPIPS on both tasks among the compared methods.
Relative to Spatia, NVS PSNR improves by 0.82 dB to 17.568 dB,
while closed-loop PSNR improves by 4.48 dB to 20.455 dB.
Closed-loop SSIM increases from 0.6827 to 0.7692, and LPIPS
falls from 0.3373 to 0.2200, showing gains in both structural
and perceptual similarity.
The larger improvements on revisits highlight the model's
ability to recover previously observed views after exploration.
This pattern aligns with the ablations in Sec.~\ref{sec:ablation},
where disabling history writing or masking the second readout
layer produces larger PSNR drops on closed-loop reconstruction
than on NVS.

\begin{figure}[t]
    \centering
    \includegraphics[width=0.74\linewidth]{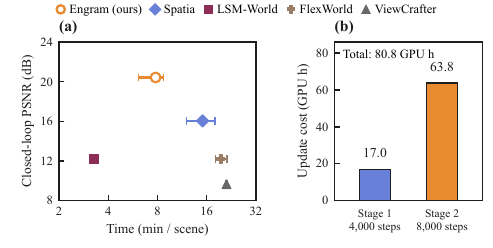}
    \caption{
        \textbf{Reconstruction quality and computational cost.}
        (a) Closed-loop PSNR versus median inference time;
        bars show runtime interquartile ranges.
        Hardware and sampling settings vary across methods.
        (b) Training-update GPU hours for the two adaptation stages.
    }
    \label{fig:efficiency}
\end{figure}

\subsection{Computational cost}
\label{sec:cost}
Figure~\ref{fig:efficiency} summarizes the recorded quality--cost
trade-off and adaptation cost.
Our two-stage adaptation uses 80.8 GPU hours of logged updates:
17.0 for Stage~1 and 63.8 for Stage~2.
Direct latent-memory readout avoids intermediate RGB rendering
and VAE encoding when constructing spatial conditions.

\begin{table}[t]
\centering
\caption{\textbf{Memory and adaptation ablations on 100 RealEstate10K scenes.}
All variants use the same scenes and evaluation protocol.
Best and second-best scores are bold and underlined, respectively.}
\label{tab:core_ablation}
\footnotesize
\setlength{\tabcolsep}{3pt}
\resizebox{0.70\linewidth}{!}{%
\begin{tabular}{@{}lcccccc@{}}
\toprule
& \multicolumn{3}{c}{NVS}
& \multicolumn{3}{c}{Closed-loop revisit} \\
\cmidrule(lr){2-4}\cmidrule(lr){5-7}
Variant
& PSNR$\uparrow$ & SSIM$\uparrow$ & LPIPS$\downarrow$
& PSNR$\uparrow$ & SSIM$\uparrow$ & LPIPS$\downarrow$ \\
\midrule
\Engram{} (full)
& \textbf{17.568} & \textbf{0.6975} & \textbf{0.3212}
& \textbf{20.455} & \textbf{0.7692} & \textbf{0.2200} \\
\midrule
No-write
& 17.052 & 0.6857 & 0.3302
& 17.284 & 0.7167 & 0.3473 \\
No-read
& 12.213 & 0.5795 & 0.5999
& 10.633 & 0.5721 & 0.7608 \\
Second-layer masked
& \underline{17.321} & \underline{0.6924} & \underline{0.3253}
& \underline{19.200} & \underline{0.7423} & \underline{0.2678} \\
Stage 1 only
& 16.777 & 0.6769 & 0.3423
& 18.570 & 0.7253 & 0.2681 \\
\bottomrule
\end{tabular}%
}
\end{table}

\begin{samepage}
\subsection{Core operator ablations}
\label{sec:ablation}
Table~\ref{tab:core_ablation} examines memory writing, memory conditioning, the second readout layer, and Stage~2 adaptation.
\par
\end{samepage}

\emph{No-write} keeps the branch, the camera conditioning and the initial-image cache, and disables only the write-back of generated observations, so memory never grows beyond the first view. This isolates what generated history contributes on top of the initial observation. NVS PSNR drops by about $0.5$ dB. Closed-loop PSNR drops by $3.2$ decibels, and LPIPS rises from $0.220$ to $0.347$. The larger drop in closed-loop reconstruction indicates a stronger dependence on generated-history writing.

\emph{No-read} removes the memory condition from the architecture entirely and is trained separately under the same recipe, with the 102 memory channels set to zero while the six camera channels are retained. NVS and closed-loop PSNR decrease to 12.213 and 10.633 dB, respectively. This highlights the importance of memory conditioning for reconstruction quality on both tasks.

\emph{Second-layer masked} zeroes the second-layer channels ($51{:}102$) of the same checkpoint and leaves the remaining operators untouched. Relative to the Full results reported in Table~\ref{tab:core_ablation}, NVS and closed-loop PSNR are lower by $0.25$ and $1.25$ dB, respectively. The original candidate-selection and initial-view reservation rules remain active.

\emph{Stage 1 only} evaluates the Stage~1 checkpoint with the pretrained backbone under the same inference configuration. Compared with the full model, NVS and closed-loop PSNR decrease by $0.79$ and $1.89$ dB, respectively, showing the benefit of Stage~2 adaptation.

\subsection{Revisiting generated content}
\label{sec:revisit_exp}
Closed-loop scoring returns to the real initial image, which can reward initial-view anchoring rather than memory. We therefore also run an $A\!\rightarrow\!B\!\rightarrow\!C\!\rightarrow\!B$ route on static development scenes, in which a first visit $B$ exposes content absent from the seed $A$ and is itself generated rather than observed. The \Engram{} model is compared with a control that keeps the initial-image cache and disables all later writes, holding camera conditioning and the generated-prefix rule fixed.

On the 20 development scenes in Table~\ref{tab:dev20}, the control reaches $17.63$ dB when returning to $A$ and $12.47$ dB when revisiting $B$, against $20.88$ and $16.87$ dB for \Engram{} with SSIM of $0.7496$ and $0.6636$. The margin is therefore larger at the generated view ($+4.4$ dB) than at the observed one ($+3.2$ dB), which is the ordering an initial-view-anchoring explanation does not predict: a model that only replays the first image has less to lose at the pose it was conditioned on. Because $B$ was itself generated, the revisit reference is the first-generated frame at that pose and not a ground-truth image, so these numbers measure persistence rather than accuracy against an unseen view, and we treat them as development-scale evidence.

\subsection{Discussion and limitations}
\label{sec:limits}

Both NVS and closed-loop evaluation use multi-chunk generation.
Removing generated-history writing or masking the second readout
layer causes a larger PSNR drop in closed-loop reconstruction
than in NVS. These results highlight the role of persistent
memory when returning to earlier viewpoints. History writing
preserves earlier observations across chunks, while layered
readout supplies candidates from multiple surfaces as visibility
changes. We report the two tasks separately to distinguish
novel-view reconstruction from revisit consistency.

\Engram{} relies on estimated geometry. Depth is predicted from decoded
frames and determines where latent observations are written in the
world coordinate system. Errors in these estimates can therefore
propagate through later memory queries, displacing stored observations
and altering the depth ordering of readout candidates.
Compatibility checks and bounded accumulation regulate memory updates,
but systematic depth bias remains a limitation. Memory grows as new
regions are explored: the per-voxel slot limit bounds local density,
not total storage. Moving surfaces are not explicitly modelled.

\section{Conclusion}
\label{sec:conclusion}
We presented \Engram{}, a latent spatial memory framework
combining reliability-weighted latent fusion and
occlusion-aware layered readout.
The framework consolidates compatible observations and exposes
multiple depth-ordered candidates within a read--generate--update cycle.
Generated-history write-back retains scene evidence beyond the
current context window, while explicit camera rays provide
viewpoint information independently of memory coverage.
Adapted with 256 training scenes, \Engram{} improves novel-view
synthesis and closed-loop reconstruction on RealEstate10K and
achieves the highest aggregate score among the compared systems
on the 200-case WorldScore subset.
The gains are particularly pronounced for closed-loop revisits.
Ablation comparisons highlight the role of generated-history
writing and layered memory conditioning in closed-loop reconstruction,
as well as the benefit of second-stage adaptation.
Future work will extend the framework to dynamic scenes and
memory management under fixed storage budgets.

\clearpage
\section*{LLM Usage Statement}
OpenAI Codex assisted with literature collection, auxiliary code development,
and manuscript editing. The authors designed the study, verified the
experimental results and all revisions, and take responsibility for all
scientific claims and interpretations.

{\small
\bibliographystyle{ieeenat_fullname}
\bibliography{engram/references}
}

\clearpage
\onecolumn
\raggedbottom
\setcounter{figure}{0}
\setcounter{table}{0}
\setcounter{equation}{0}
\renewcommand{\thefigure}{S\arabic{figure}}
\renewcommand{\thetable}{S\arabic{table}}
\renewcommand{\theequation}{S\arabic{equation}}
\renewcommand{\theHfigure}{supp.\arabic{figure}}
\renewcommand{\theHtable}{supp.\arabic{table}}
\renewcommand{\theHequation}{supp.\arabic{equation}}
\makeatletter
\setlength{\@fptop}{0pt}
\setlength{\@fpsep}{16pt}
\setlength{\@fpbot}{0pt plus 1fil}
\makeatother
\begin{center}
    {\Large\bfseries \thetitle\par}
    \vspace{0.5em}
    {\large Supplementary Material\par}
\end{center}
\vspace{0.5em}
\appendix
\section{Analysis of the Weighted Fusion Rule}
\label{app:theory}

We analyze repeated observations of a fixed scene location in the normalized latent coordinates of Sec.~\ref{sec:fusion}. Let $\ff_i=\ff^\star+\bfeps_i$, $i=1,\dots,n$, where $\ff^\star\in\real^C$ is the unknown latent feature. For the independent-noise analysis, we assume
$\Exp[\bfeps_i]=0$,
$\Covv(\bfeps_i)=\sigma_i^2 I$,
$0<\sigma_i^2<\infty$,
and mutually independent errors. These are assumptions of the observation model. The correlated case is considered separately below.

\subsection{Inverse-variance weighting}
\label{app:weightedmean}

Consider one scalar channel and a linear estimator $\hat f=\sum_i c_i f_i$ with fixed coefficients. Unbiasedness for every $f^\star$ requires $\sum_i c_i=1$. Under independence,
\begin{equation}
\Varr(\hat f)=\sum_i c_i^2\sigma_i^2.
\end{equation}
Minimizing this variance subject to $\sum_i c_i=1$ with a Lagrange multiplier $\lambda$ gives $2c_i\sigma_i^2=\lambda$. Hence the minimizing coefficients are
\begin{equation}
c_i^\star
=
\frac{\sigma_i^{-2}}{\sum_j\sigma_j^{-2}}.
\end{equation}
Let $\omega_i=\sigma_i^{-2}$ denote the inverse variance. Applying the same coefficients to all latent channels gives
\begin{equation}
\begin{aligned}
\hat{\ff}
&=
\frac{\sum_i\omega_i\ff_i}{\sum_i\omega_i},\\
\Covv(\hat{\ff})
&=
\Bigl(\sum_i\sigma_i^{-2}\Bigr)^{-1}I
\preceq
\min_i\sigma_i^2\,I.
\end{aligned}
\label{eq:ideal_fusion}
\end{equation}
The estimator is unbiased because its coefficients sum to one. The covariance inequality is strict for $n\ge2$ under the positive, finite variance assumptions above. This establishes the best linear unbiased estimator under the stated model~\citep{casella2002statistical}.

\subsection{Uncapped online accumulation}

Let $S_t=\sum_{i=1}^{t}\omega_i$ and $\hat{\ff}_t=S_t^{-1}\sum_{i=1}^{t}\omega_i\ff_i$. For $t\ge2$,
\begin{equation}
\hat{\ff}_t
=
\frac{S_{t-1}\hat{\ff}_{t-1}+\omega_t\ff_t}
     {S_{t-1}+\omega_t}.
\label{eq:ideal_online}
\end{equation}
Thus, uncapped accumulation equals the batch estimator. Permuting the same observation--weight pairs leaves the estimate unchanged because its numerator and denominator are sums. Writing $\alpha_t=\omega_t/S_t$ gives $\hat{\ff}_t=(1-\alpha_t)\hat{\ff}_{t-1}+\alpha_t\ff_t$.

Under inverse-variance weighting, $\Covv(\hat{\ff}_t)=S_t^{-1}I$. This covariance tends to zero when $\sum_i\omega_i=\infty$. A sufficient condition is $\omega_i\ge\omega_{\min}>0$ for all observations.

The practical update in Eq.~\eqref{eq:fuse} caps the stored weight at $W_{\max}$. Once the cap alters the stored weight, the batch-estimator identity and order independence generally no longer hold. The limiting behavior then depends on the incoming observations and their weights.

\subsection{Single-observation selection}

For an index $j$ fixed in advance, selecting $\ff_j$ gives an unbiased estimator with covariance $\sigma_j^2 I$. Equation~\eqref{eq:ideal_fusion} therefore gives
\begin{equation}
\Covv(\hat{\ff})
\preceq
\Covv(\ff_j),
\end{equation} with strict inequality for $n\ge2$ under the stated assumptions.

Depth-based selection instead uses an index determined by the observed depths. For repeated measurements of the same surface, let $d_i=d^\star+\eta_i$ and $J=\arg\min_i d_i$. Assume $n\ge2$ and i.i.d., integrable, non-degenerate depth errors with $\Exp[\eta_i]=0$.
Then
\begin{equation}
\begin{aligned}
\Exp[d_J-d^\star]
&=\Exp\bigl[\min_i\eta_i\bigr]\\
&\le\Exp\bigl[\min(\eta_1,\eta_2)\bigr]\\
&=-\frac{1}{2}\Exp\bigl[|\eta_1-\eta_2|\bigr]
<0.
\end{aligned}
\end{equation}
Thus, minimum-depth selection biases the selected depth toward the camera under this model.

The bias of the associated latent feature is determined by
\begin{equation}
\Exp[\ff_J-\ff^\star]
=
\sum_j
\Pr(J=j)\,
\Exp[\bfeps_j\mid J=j].
\end{equation}
When depth ordering is independent of all latent errors, this expectation is zero. Latent-feature bias therefore depends on the relationship between the selection event and the feature errors.

\subsection{Correlation and imperfect weights}

For one scalar channel, suppose
$\Varr(\varepsilon_i)=\sigma^2$ and
$\Covv(\varepsilon_i,\varepsilon_j)=\rho\sigma^2$
for $i\ne j$, with $\rho\in[0,1)$.
The equally weighted mean has variance
\begin{equation}
\begin{aligned}
\Varr(\hat f)
&=
\frac{1}{n^2}
\Bigl(n\sigma^2+n(n-1)\rho\sigma^2\Bigr)\\
&=
\frac{\sigma^2}{n}\bigl(1+(n-1)\rho\bigr)
=
\frac{\sigma^2}{n_{\mathrm{eff}}},
\end{aligned}
\end{equation}
where
\begin{equation}
n_{\mathrm{eff}}
=
\frac{n}{1+(n-1)\rho}.
\end{equation}
For $\rho>0$ and $n>1$, $n_{\mathrm{eff}}<n$. As $n\to\infty$, the effective sample size approaches $1/\rho$ and the variance approaches $\rho\sigma^2$.

More generally, let $V$ be the covariance matrix of the scalar observation errors. For fixed coefficients $\mathbf c$ satisfying
$\mathbf 1^\top\mathbf c=1$,
\begin{equation}
\Exp[\hat f]=f^\star,
\qquad
\Varr(\hat f)=\mathbf c^\top V\mathbf c.
\end{equation}
Under independence, this reduces to $\sum_i c_i^2\sigma_i^2$. Fixed weights that differ from inverse variances remain unbiased under the zero-mean model, but need not minimize the variance.

For observation-dependent coefficients, unbiasedness requires
\begin{equation}
\Exp\Bigl[\sum_i c_i\varepsilon_i\Bigr]=0.
\end{equation}

\subsection{Affine normalization and weighted averaging}

Let $g(\ff)=\Sigma^{-1/2}(\ff-\boldsymbol{\mu})$, with fixed $\boldsymbol{\mu}$ and positive diagonal $\Sigma$. For the same observations and coefficients satisfying $\sum_i c_i=1$,
\begin{equation}
\begin{aligned}
\sum_i c_i g(\ff_i)
&=
\Sigma^{-1/2}
\Bigl(\sum_i c_i\ff_i-\boldsymbol{\mu}\Bigr)\\
&=
g\Bigl(\sum_i c_i\ff_i\Bigr).
\end{aligned}
\end{equation}
Weighted averaging therefore commutes with this affine map. Applying
$g^{-1}(\tilde{\ff})=\Sigma^{1/2}\tilde{\ff}
+\boldsymbol{\mu}$ restores the corresponding average in the original latent coordinates.

This identity holds for the same observations and weights. In the memory operator, normalization also determines the feature distances used for compatibility testing. Its role is to place channels on comparable scales; the noise assumptions used above are specified separately.

\section{Implementation and Training Details}
\label{app:details}

\paragraph{Memory configuration.}
The latent has $C=48$ channels, and each voxel accommodates at most $L=8$ slots. The compatibility thresholds are $\tau_{\mathrm f}=3$ and $\tau_{\mathrm p}=1$, with position distances divided by the voxel side length. The accumulated weight is capped at $W_{\max}=20$. Scene coordinates are normalized by the median metric depth $s_0$ of the initial image. A voxel side length of $0.005$ metres therefore becomes $h_{\mathrm{vox}}=(0.005\,\mathrm{m})/s_0$ in normalized coordinates. A generated slot adopts the incoming candidate position only when the candidate weight exceeds the stored weight before the update.

\paragraph{Normalization and reliability.}
Channel normalization uses fixed statistics and a variance stabilizer of $10^{-5}$. The angular factor in Eq.~\eqref{eq:reliability} uses $\epsilon=10^{-3}$, and the temporal discount base is $\gamma=0.98$. The angular exponent is $1/2$, and the temporal factor is $\gamma^{r_i}$. Here, $r_i=\lfloor s_i/4\rfloor$, where $s_i$ is the global RGB-frame index measured from the rollout start. Initial-observation weights are multiplied by four. Generated observations are fused into separate slots, while the initial-observation slots remain fixed.

\paragraph{Layered conditioning.}
The readout returns up to $K=2$ layers. Each layer contains $C$ feature channels, one reliability channel, one validity mask, and one inverse-depth channel. The memory representation therefore has $102$ channels; appending the six Pl\"ucker ray channels gives $108$. The reliability channel is
\begin{equation}
q_a
=
\frac{\min(W_a,5)}{1+\min(W_a,5)}.
\end{equation}
The readout cap of five is applied separately from the fusion weight cap $W_{\max}=20$. The inverse-depth channel is $1/(1+d)$, where $d$ is the positive camera depth in scene-normalized coordinates. Missing layers are filled with zeros and marked invalid. The conditioning branch uses blocks copied from backbone layers $\{0,4,8,12,16,20,24,28\}$.

\paragraph{Generation and memory updates.}
Each chunk contains nine latent frames corresponding to 33 RGB frames. Consecutive chunks overlap by one RGB frame, giving a stride of 32 frames. Sampling uses UniPC~\citep{zhao2023unipc} with 40 steps. Depth is estimated from decoded frames using MapAnything~\citep{keetha2026mapanything}. Writes require finite positive depth. A candidate projecting onto valid initial-view depth is rejected when its depth in that camera is below $0.9$ times the reference depth. At each chunk boundary, the latent refinement in Eq.~\eqref{eq:handoff} uses Adam for 50 steps with learning rate $0.02$ and mean absolute RGB reconstruction error.

\paragraph{Optimization details.}
Both training stages use a learning rate of $5\times10^{-6}$.
The learning rate is constant in Stage~1; Stage~2 uses a
100-step linear warm-up followed by a constant learning rate.

\paragraph{Recorded training cost.}
The logged training-update costs are 17.0 GPU hours for Stage~1
and 63.8 GPU hours for Stage~2, totaling 80.8 GPU hours.
These costs exclude data preparation, initialization,
and checkpoint saving.

\subsection{Training-sample accounting}
\label{sec:training_accounting}
We count training-sample presentations as
$P=\sum_s T_s B_s$, where $T_s$ is the number of optimization
steps and $B_s$ is the effective batch size in stage $s$.
Repeated draws of the same sample each count as a presentation.
For \Engram{}, both stages use an effective batch size of four:
\[
P_{\mathrm{Engram}}=4{,}000\times4+8{,}000\times4=48{,}000.
\]
Spatia reports 8,000 ControlNet updates followed by 5,000 LoRA
updates, with a batch size of 64 in both stages~\citep{zhao2026spatia}.
Interpreting the reported batch size as the effective batch size gives
\[
P_{\mathrm{Spatia}}=8{,}000\times64+5{,}000\times64=832{,}000.
\]
Thus, $P_{\mathrm{Engram}}/P_{\mathrm{Spatia}}\approx5.8\%$,
and the reciprocal is approximately $17.3$.
Spatia's 50,000 source videos describe the dataset size and do not
enter this ratio. These are schedule-derived presentation counts;
sample duration, resolution, and trainable parameters differ between
methods, so the ratio does not measure unique-data requirements,
processed frames or tokens, GPU hours, or computational speedup.

\section{Evaluation and Ablation Protocols}
\label{app:ablation}

\subsection{Benchmark evaluation protocols}
\label{sec:eval_protocol}

RealEstate10K evaluates reconstruction fidelity, while
WorldScore provides complementary measures of camera control
and motion quality.

\paragraph{WorldScore.}
Table~\ref{tab:worldscore} reports the ten official metrics and the WorldScore-Static and WorldScore-Dynamic aggregates~\citep{duan2025worldscore}, with Avg.\ defined as their mean. Static averages seven control and quality metrics; Dynamic averages those seven plus three motion metrics. These aggregates are distinct from the static and dynamic sample counts. WonderWorld generates static scenes only, so its motion metrics and the aggregates that depend on them are not applicable and are left blank rather than filled with the benchmark's placeholder zeros. General video models receive image and text conditioning with benchmark camera prompts, whereas camera-controlled systems receive numeric poses; this input difference is constant across their rows.

\paragraph{RealEstate10K.}
Table~\ref{tab:re10k} evaluates novel-view synthesis (NVS) and closed-loop return on a common held-out scene list of 100 scenes~\citep{zhou2018stereo,zhao2026spatia}. NVS generates 81 frames and scores frames 1--80 against their ground-truth counterparts. Closed-loop generation follows a fixed 153-pose trajectory and scores frame 152 against the initial image. Each task starts independently from the same real image and prescribed camera annotations. Subsequent visual history contains only generated observations; held-out RGB frames are used solely for scoring, including when adapting camera-input interfaces. All methods generate the full evaluation trajectories. Target field of view, camera conventions, frame times, and scoring resolution are matched across methods. NVS scores are averaged within each scene and then across scenes. We use a fixed seed policy and no per-metric checkpoint selection.

\subsection{Ablation settings}

Table~\ref{tab:core_ablation} includes inference interventions,
a separately trained No-read model, and a Stage~1 checkpoint comparison.

\paragraph{Inference interventions.}
\emph{No-write} and \emph{second-layer masked} reuse the same \Engram{}
checkpoint as the full row and change only the runtime path. No-write retains the
initial-image cache, the camera conditioning and the generated-prefix rule, and
suppresses the write-back of generated observations, so the memory never grows
beyond the first view. Second-layer masked zeroes the second-layer conditioning
channels ($51{:}102$) while leaving selection depth $K=2$, the first layer, the
camera channels and the reservation of real initial-view candidates intact. It is
therefore not a retrained single-layer model, and the residual gap it produces
combines the second layer with the reservation rule rather than isolating either.
Because both rows reuse the full checkpoint, their differences from the full
model are attributable to the removed component without retraining variance, but
they cannot be read as the performance of a system designed without that
component.

\paragraph{Retrained variant.}
\emph{No-read} removes the memory condition from the architecture and is trained
separately under the same data, schedule and objective, with the 102 memory
channels set to zero and the six camera channels retained.

\paragraph{Stage~2 adaptation.}
\emph{Stage 1 only} loads the Stage~1 side-branch checkpoint on the unmodified
pretrained backbone, retaining the full model's memory operators,
inference path, hyperparameters and score normalization.
This comparison evaluates the benefit of Stage~2 adaptation.

\paragraph{Sharing.}
All five rows use the same 100 test scenes, camera trajectories, target
resolution, sampler and scoring code. Scenes are scored individually and then
averaged, so no row benefits from a different subset.

\clearpage
\section{Generated-Content Revisit}
\label{app:revisit}

\begin{figure}[H]
\centering
\includegraphics[width=\linewidth]{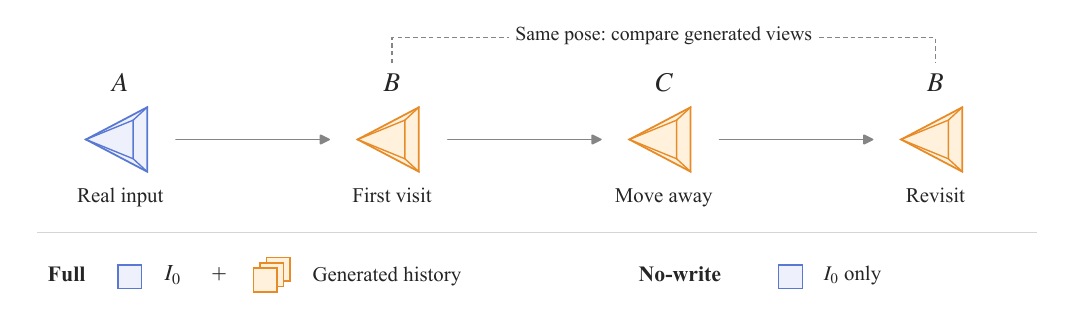}
\caption{Generated-content revisit protocol. View $B$ reveals content absent
from the seed $A$ and is generated rather than observed, so the return at $B$
is scored against the first-generated frame at that pose and not against
ground truth. Camera conditioning and the generated-prefix continuation rule
are held fixed between the two rows.}
\label{fig:generated_revisit}
\end{figure}

Closed-loop scoring in Table~\ref{tab:re10k} returns to the \emph{real} initial
image, which can reward initial-view anchoring rather than memory. We therefore
also evaluate a route in which the return target is itself generated
(Fig.~\ref{fig:generated_revisit}): starting from a real seed $A$, the model
generates a first visit $B$ that exposes content absent from $A$, leaves toward
$C$, and then returns to the pose of $B$.

The comparison holds camera conditioning and the generated-prefix rule fixed and
differs only in whether generated observations are written back. Because $B$ was
never observed, the revisit reference is the first-generated frame at that pose
and not a ground-truth image, so the revisit column measures persistence of
generated content rather than accuracy against an unseen view. The return to $A$
is scored against the real input image and does measure accuracy.

\begin{table}[H]
\centering
\begin{minipage}{0.78\linewidth}
\centering
\caption{\textbf{Development-scale revisit comparison.}
Evaluation on 20 static development scenes, disjoint from test-100.
Return to $A$ uses the real input image as reference; revisit $B$ uses the
first-generated frame at the same pose (Fig.~\ref{fig:generated_revisit}).
No-write retains the initial-image cache and disables later writes.
Both variants share the camera trajectories, sampler, and scoring code.}
\label{tab:dev20}
\small
\setlength{\tabcolsep}{4pt}
\begin{tabular}{@{}lcccccc@{}}
\toprule
& \multicolumn{3}{c}{Return to $A$}
& \multicolumn{3}{c}{Revisit $B$} \\
\cmidrule(lr){2-4}\cmidrule(lr){5-7}
Variant & PSNR$\uparrow$ & SSIM$\uparrow$ & LPIPS$\downarrow$
& PSNR$\uparrow$ & SSIM$\uparrow$ & LPIPS$\downarrow$ \\
\midrule
\Engram{} (full) & 20.876 & 0.7496 & 0.2131 & 16.866 & 0.6636 & 0.4799 \\
No-write & 17.629 & 0.6921 & 0.3275 & 12.465 & 0.5172 & 0.6971 \\
\bottomrule
\end{tabular}
\end{minipage}
\end{table}

Table~\ref{tab:dev20} reports both columns on 20 static development scenes held
out from the test-100 evaluation. The margin is larger at the generated view than
at the observed one. A model that only replays the initial image has less to lose
at the pose it was conditioned on, so this ordering is not what initial-view
anchoring predicts. We treat these scenes as development-scale evidence: they are
fewer than the test-100 set and were not reserved for a final comparison.

\clearpage
\section{Qualitative Results}
\label{app:qualitative}

This section complements Tables~\ref{tab:re10k} and~\ref{tab:worldscore} with
archived rollouts. Images are uniformly resized without enhancement or retouching.
LSM-World results use our reproduction, as no checkpoint was released.

\subsection{Closed-loop revisit}
\label{app:qual_closedloop}
Figure~\ref{fig:qual_closedloop} compares ViewCrafter, FlexWorld, LSM-World,
Spatia, and \Engram{} on the same four scenes. The camera leaves the input view
and returns; each method's frame 152 is compared with the real input image.
The PSNR labels report individual scene scores rather than dataset averages.

\begin{figure}[H]
    \centering
    \includegraphics[width=\textwidth]{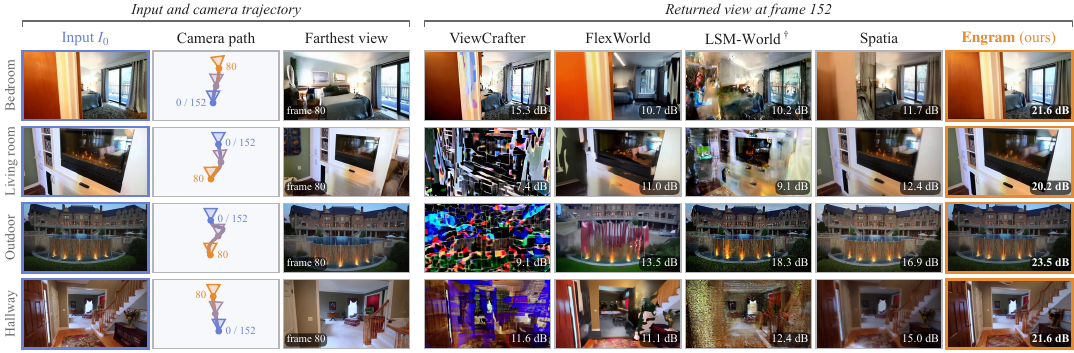}
    \input{engram/figure_text/qual_captions_closedloop}
\end{figure}

\clearpage
\subsection{Novel-view synthesis}
\label{app:qual_nvs}
Figure~\ref{fig:qual_nvs} follows the target camera trajectory through four
RealEstate10K scenes. Ground-truth frames and the five methods are aligned at
eight frames, from 10 to 80 in steps of 10. Each method's PSNR is averaged over frames 1--80
of that scene, using the same scoring protocol as Table~\ref{tab:re10k}.

\begin{figure}[H]
    \centering
    \includegraphics[width=\textwidth]{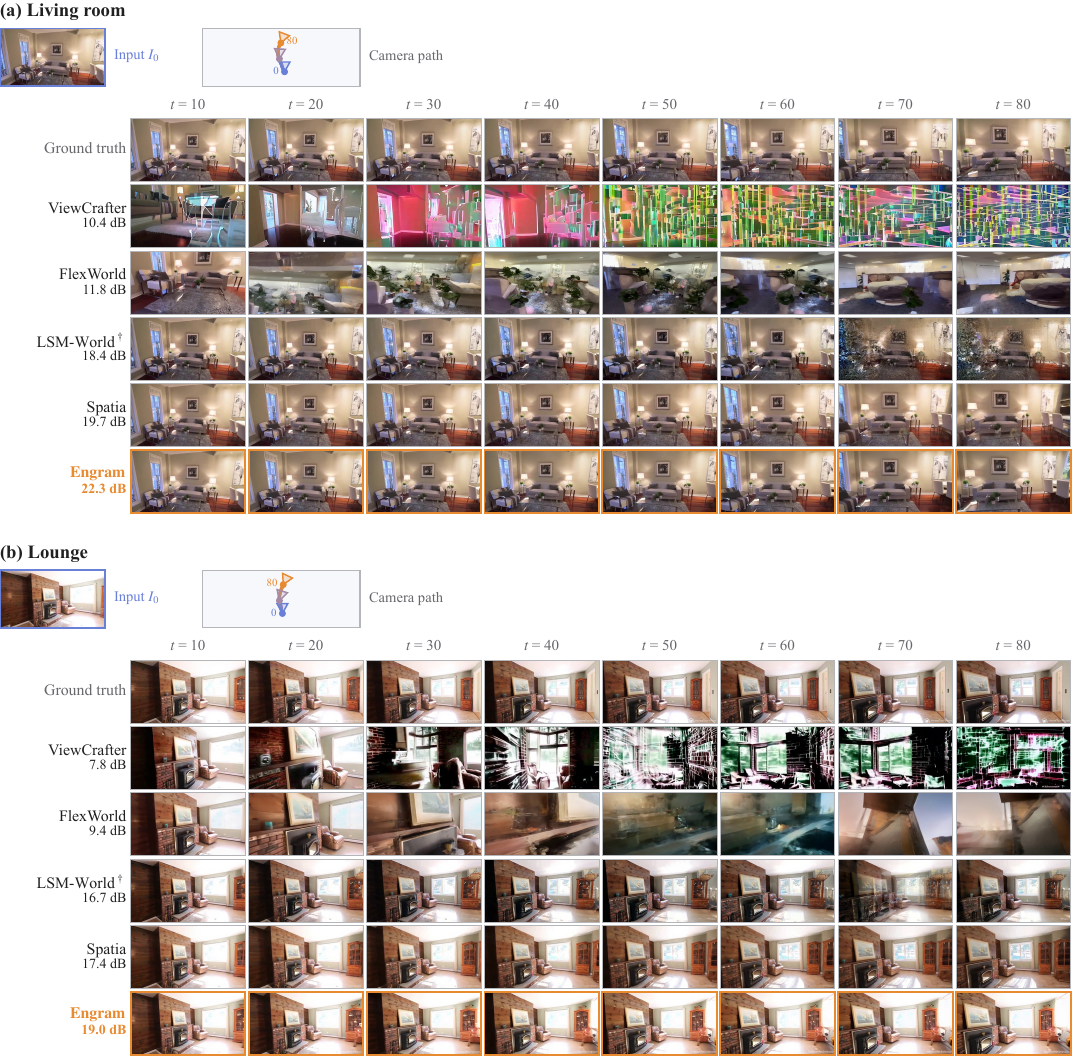}
    \input{engram/figure_text/qual_captions_nvs}
\end{figure}

\clearpage
\begin{figure}[H]
    \ContinuedFloat
    \centering
    \includegraphics[width=\textwidth]{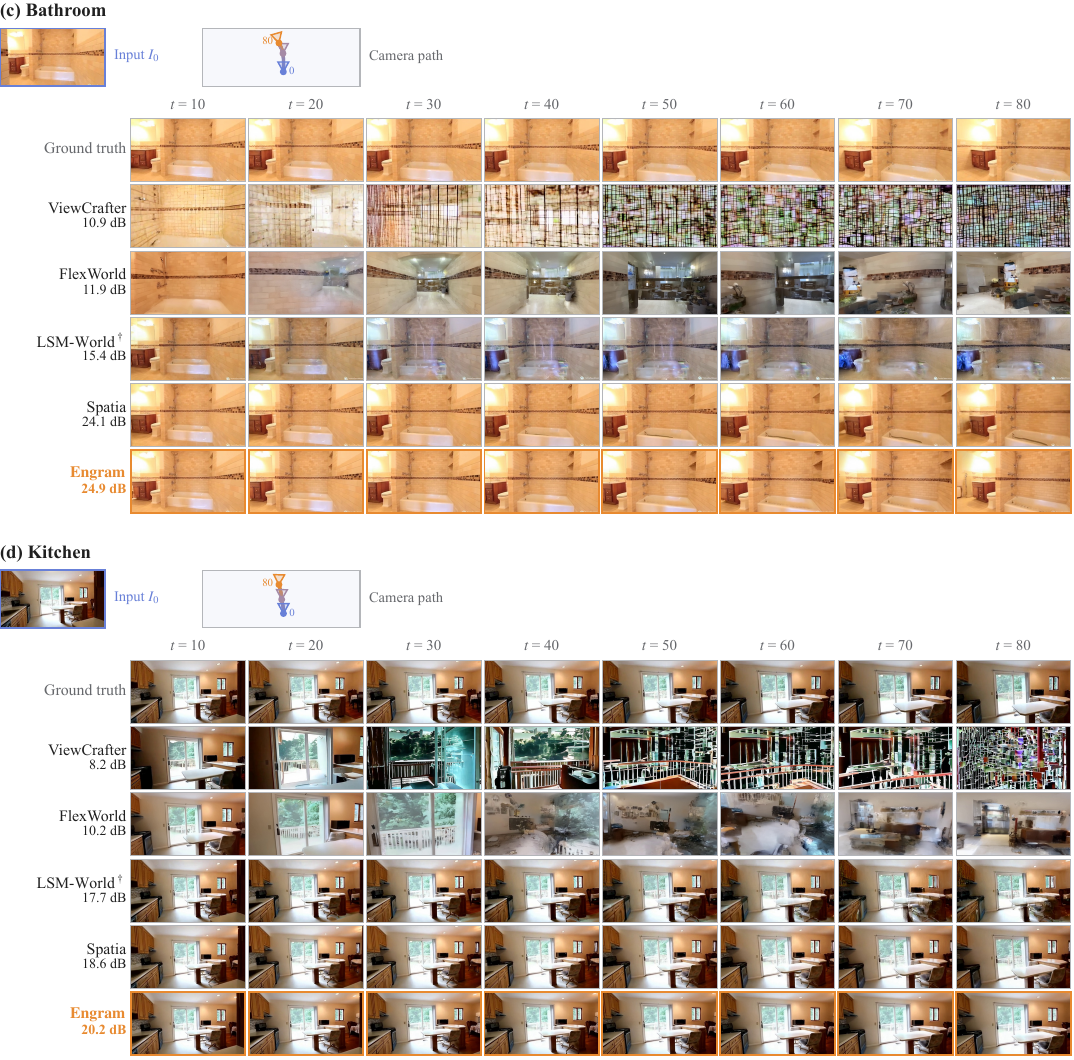}
    \caption{\textbf{Novel-view synthesis (continued).}
    (c) Bathroom and (d) kitchen, using the same methods, frame indices,
    and scoring protocol as panels (a)--(b).
    $^\dagger$Our reproduction.}
\end{figure}

\clearpage
\subsection{WorldScore generation}
\label{app:qual_worldscore}
Figure~\ref{fig:qual_worldscore} compares Wan2.2-TI2V-5B, CogVideoX,
LTX-Video, Spatia, LSM-World, and \Engram{} on five benchmark inputs.
The scenes span photorealistic and stylized, indoor and outdoor environments,
and a dynamic case. For each scene, two aligned rows show how the six rollouts
evolve under the indicated camera command.

\begin{figure}[H]
    \centering
    \includegraphics[width=\textwidth]{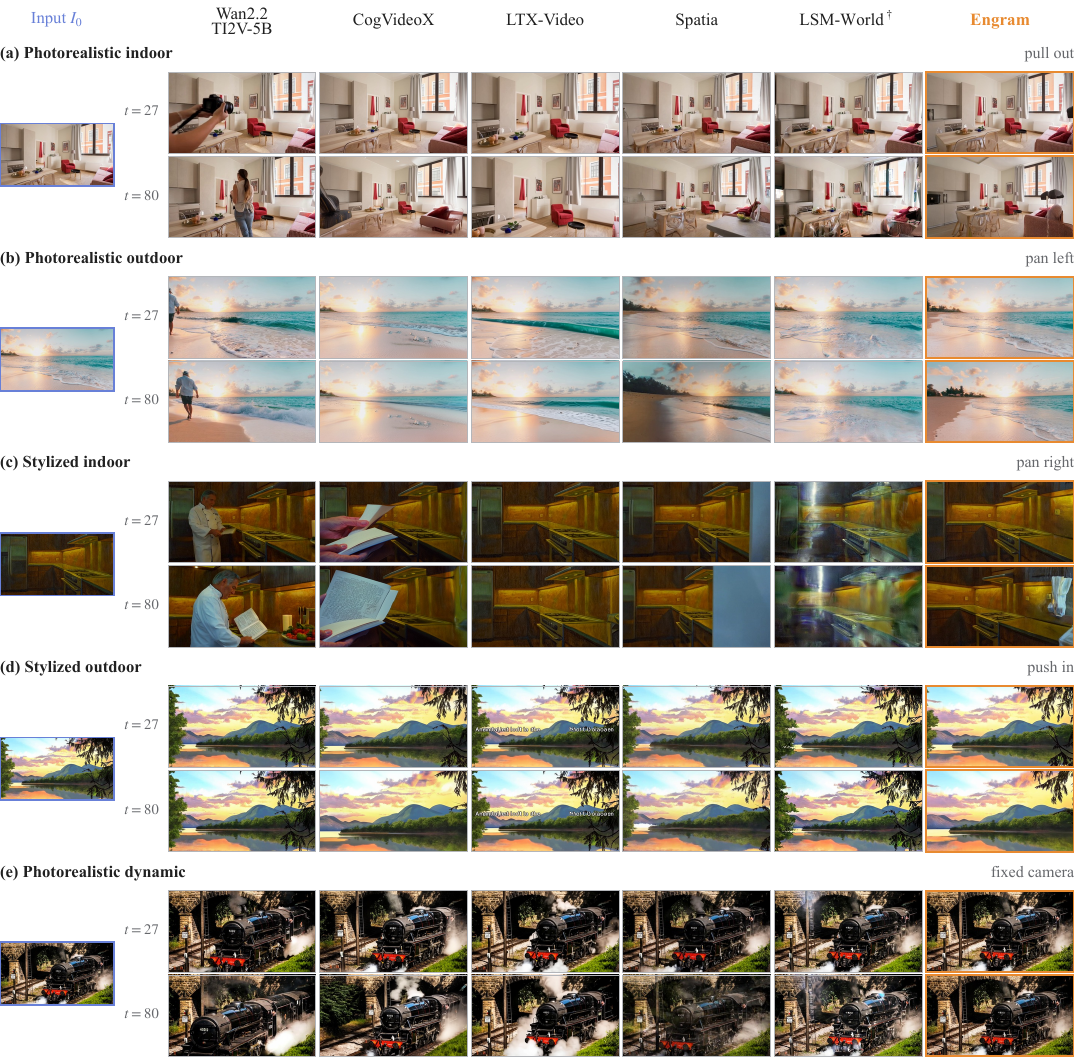}
    \input{engram/figure_text/qual_captions_worldscore}
\end{figure}

\clearpage
\subsection{Failure cases}
\label{app:qual_failure}
Figure~\ref{fig:qual_failure} shows two limitations on the same RealEstate10K
test protocol. In the first case, newly revealed content differs from the
ground truth; in the second, an unsupported dark occluder appears late in the
rollout. The ground truth and all five methods are shown at matching frames.

\begin{figure}[H]
    \centering
    \includegraphics[width=\textwidth]{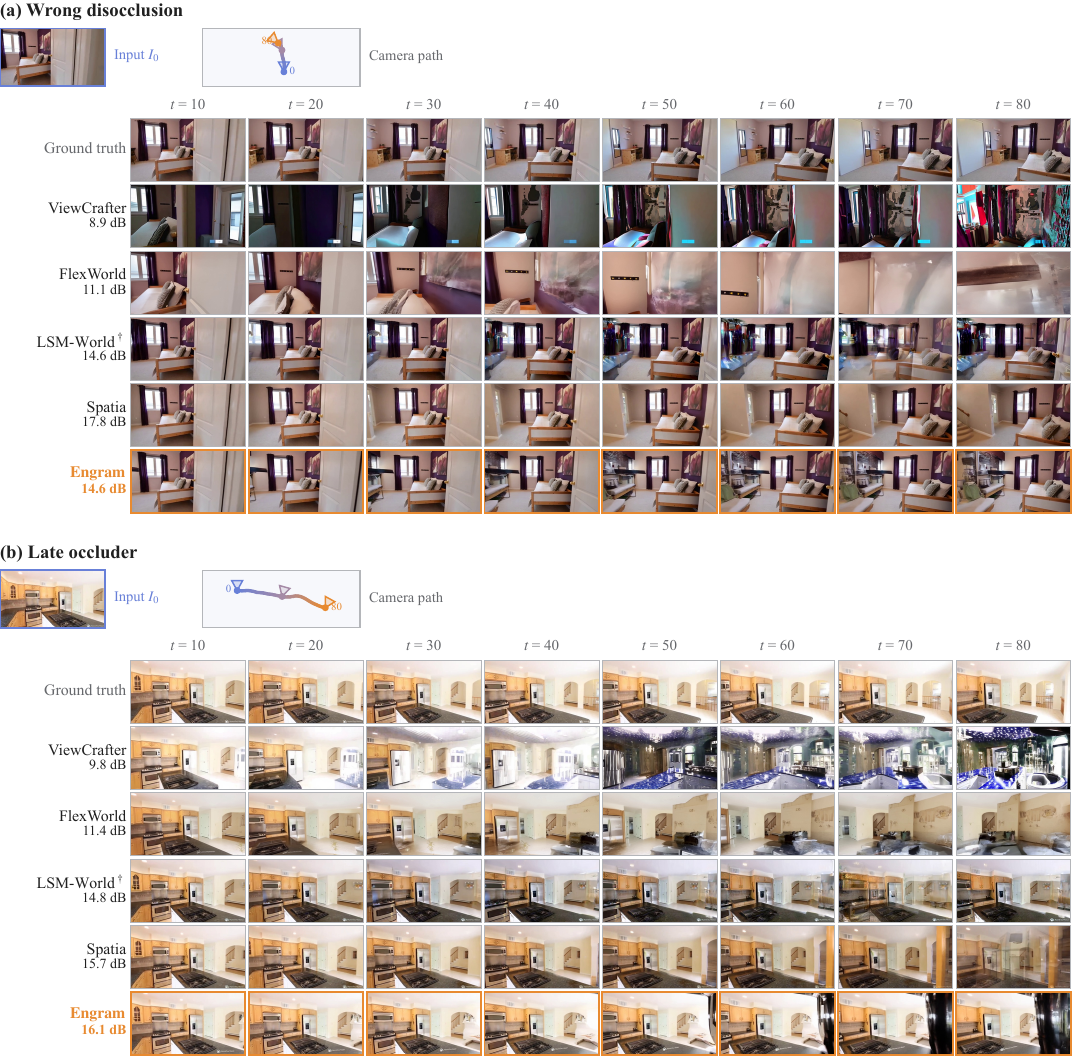}
    \input{engram/figure_text/qual_captions_failure}
\end{figure}

\end{document}